\documentclass{bmvc2k}

\usepackage{graphicx}
\usepackage{tikz}
\usepackage{comment}
\usepackage{amsmath,amssymb} 
\usepackage{color}
\usepackage{epstopdf}
\usepackage{wrapfig}
\usepackage{booktabs}
\usepackage{multirow}
\usepackage[accsupp]{axessibility}

\makeatletter
\newcommand\figcaption{\def\@captype{figure}\caption}
\newcommand\tabcaption{\def\@captype{table}\caption}
\makeatother

\title{LW-ISP: A \underline{L}ight\underline{w}eight Model with \underline{ISP} and Deep Learning}

\addauthor{Hongyang Chen}{chenhy@stu.xjtu.edu.cn}{1}
\addauthor{Kaisheng Ma}{kaisheng@mail.tsinghua.edu.cn}{2}

\addinstitution{
 Xi’an Jiaotong University\\
 Xi'an, China
}
\addinstitution{
Tsinghua University\\
Beijing, China
}

\runninghead{Chen, Ma}{LW-ISP: A Lightweight Model}

\begin{document}
\maketitle


\begin{abstract}
The deep learning (DL)-based methods of low-level tasks have many advantages over the traditional camera in terms of hardware prospects, error accumulation and imaging effects. Recently, the application of deep learning to replace the image signal processing (ISP) pipeline has appeared one after another; however, there is still a long way to go towards real landing.
In this paper, we show the possibility of learning-based method to achieve real-time high-performance processing in the ISP pipeline. We propose \texttt{LW-ISP}, a novel architecture designed to implicitly learn the image mapping from RAW data to RGB image. Based on U-Net architecture, we propose the fine-grained attention module and a plug-and-play upsampling block suitable for low-level tasks. In particular, we design a heterogeneous distillation algorithm to distill the implicit features and reconstruction information of the clean image, so as to guide the learning of the student model.
Our experiments demonstrate that \texttt{LW-ISP} has achieved a 0.38 dB improvement in PSNR compared to the previous best method, while the model parameters and calculation have been reduced by 23$\times$ and 81$\times$. The inference efficiency has been accelerated by at least 15$\times$. Without bells and whistles, \texttt{LW-ISP} has achieved quite competitive results in ISP subtasks including image denoising and enhancement.
\end{abstract}

\section{Introduction}
\label{intro}

 In recent years, smartphones have increasingly dominated daily photos. With the emergence of advanced applications such as autonomous driving~\cite{van2018autonomous,caesar2020nuscenes}, high-speed continuous shooting~\cite{zhan2020hdr} and 4K recording~\cite{chandrappa2017use}, the importance and requirements for cameras are increasing gradually. The image signal processing (ISP)~\cite{nishimura1987three,ramanath2005color,heide2014flexisp,wu2019visionisp} is used to receive and process the raw signal of the sensor during the entire process of camera imaging, which has a decisive effect on the quality of the image. As mobile devices begin to adapt to powerful hardware with an ISP system~\cite{faggiani2014smartphone}, the resolution has been greatly increased. However, small sensors and relatively compact lenses have led to the loss of detail and high noise levels, and the current ISP system still fails to solve these problems completely. Moreover, as Deep Neural Networks (DNNs) achieve performance that surpasses conventional algorithms in the tasks of image classification~\cite{wang2017residual,chang2020devil}, speech recognition~\cite{han2020contextnet,gulati2020conformer} and other fields~\cite{tan2020efficientdet,minaee2020image,braso2020learning}, the combination of DNNs and ISP has been brought to the fore.
 
\begin{wrapfigure}{r}{0.48\textwidth}
\setlength{\abovecaptionskip}{-15pt}
\setlength{\belowcaptionskip}{-15pt}
\centering
\includegraphics[width=0.95\linewidth]{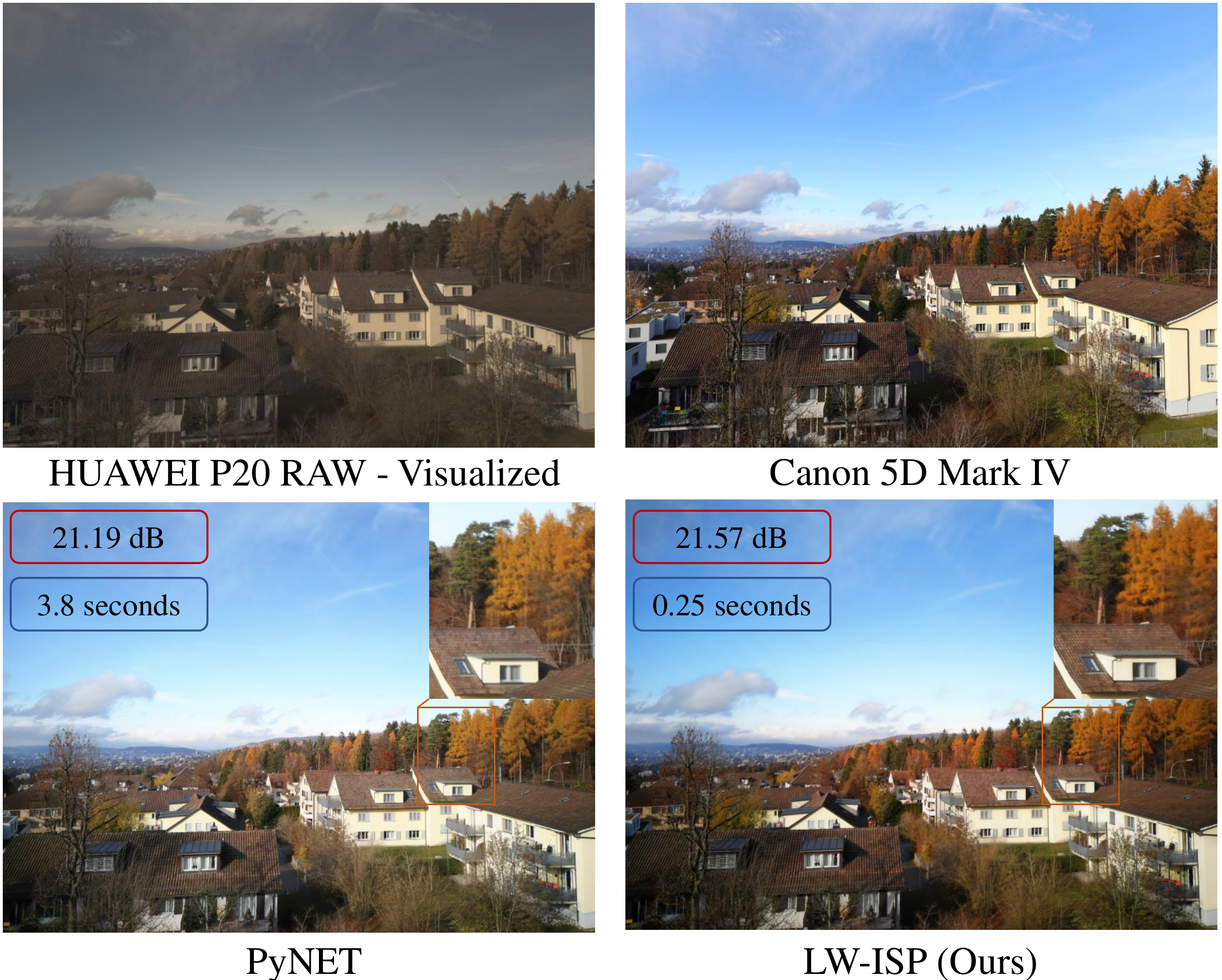}
\caption{Example set of images from ISP dataset (Zurich~\cite{ignatov2020replacing}). The number in the red box represents the PSNR value, and the blue box indicates the test time under a single NVIDIA Tesla V100.}
\label{demonstration}
\end{wrapfigure}

 Traditional ISPs and DL-based solutions still face various challenges. As the special hardware in the camera dedicated to image processing tasks, ISP can solve many low-level and global image processing tasks in proper order, such as demosaicing, white balance, exposure correction, and gamma correction. 
 In the design of the traditional ISP system, aforementioned tasks are well researched independently, without considering its subsequent impact, which may lead to the accumulation of errors in the entire processing pipeline. That means that the overall process will be affected by error propagation from stage to stage. For instance, early demosaicing artifacts may be amplified by image sharpening or misalignment of different exposures~\cite{heide2014flexisp}. 
 At present, learning ISP pipeline promotes a novel direction of research aiming at replacing the current tedious and expensive handcrafted ISP solutions with data-driven learned ones capable of surpassing them in terms of image quality. The advantages of learning-based methods are that they can implicitly learn the statistical information of natural images and allow joint solutions for multiple tasks. However, the limited research~\cite{schwartz2018deepisp,ratnasingam2019deep,hsyu2021csanet,ignatov2020replacing} mainly focuses on the improvement of objective indicators or only targets closely related tasks. These methods customarily require a higher computational overhead, which is challenging to be taken into account by the application.

 In this paper, we propose \texttt{LW-ISP} to replace the entire ISP pipeline, achieving an effective balance of processing efficiency and performance. The first step is to design a tiny U-Net as the base model and the \underline{F}ine-\underline{g}rained \underline{A}ttention \underline{M}odule (\textbf{FGAM}) to reconstruct the overall information during down-sampling processes. The second step towards incorporating contextual information into the upsampling blocks to preserve realistic details from RAW inputs, we design a plug-and-play \underline{C}ontextual \underline{C}omplement Upsampling \underline{B}lock (\textbf{CCB}). 
 Finally, we design the heterogeneous distillation algorithm to train the teacher model based on the target data and then distill the clean features from the teacher model to the student.

 Our experiments demonstrate the superiority of the \texttt{LW-ISP} on the ISP dataset and its subtasks. In a large-scale learning setup, our approach achieves a performance exceeding SOTA methods on the ISP's largest dataset and dramatically speeds up the inference process (model parameters are reduced by 23 times and calculations are reduced by 81 times), as shown in Fig.~\ref{demonstration}. Furthermore, our training techniques improve transfer performance on a suite of ISP downstream subtasks such as image denoising and image enhancement. We recommend that practitioners use this simple architecture as a baseline for future research.

 To sum up, the contribution of this paper can be summarized as follows.
 \begin{itemize}
 \item We design a novel lightweight \texttt{LW-ISP} model to replace the entire ISP pipeline. We evaluate the performance of \texttt{LW-ISP} on the ISP pipeline and different subtasks. Plainly speaking, \texttt{LW-ISP} not only outputs processed images with higher subjective and objective quality (PSNR: 0.38dB), but also takes less inference time (15$\times$).

 \item We propose a fine-grained attention module to reconstruct the overall information and determine a more reasonable upsampling block in low-level image processing.

 \item We propose a heterogeneous distillation training algorithm to distill the spatial structure information and global information of the environment from the teacher model to the student model.
 
 \end{itemize}

\section{Related Work}
\label{related work}

 \textbf{DL-based ISP Subtasks.}
 Deep learning (DL)-based methods have achieved considerable success in image preprocessing subtasks, including demosaicing~\cite{liu2020joint}, denoising~\cite{cheng2021nbnet}, deblurring~\cite{zhang2019deep} and super-resolution~\cite{mei2021image}, which all have achieved performance beyond conventional algorithms. Studies have shown that even when operating outside of a supervised learning mechanism, DNNs are proficient in generating high-quality images~\cite{ulyanov2018deep}. 
 Recently, researchers denoise on RAW images in order to avoid the effect of ISP~\cite{wei2020physics}. Brooks \textit{et al.} creatively inversely transformed the color image and used it for the training~\cite{brooks2019unprocessing}. 
 Contrary to the conventional methods of independently solving ISP subtasks, DL-based methods allow multiple tasks to be jointly solved, which has great potential to reduce the computational burden. However, existing schemes require more calculations.

 \noindent
 \textbf{DL-based ISP Pipeline.}
 The application of DNNs to solve the ISP pipeline has gradually attracted attention. As the first attempt of ISP with DL, DeepISP~\cite{schwartz2018deepisp} divided the framework into high-dimensional and low-dimensional feature extraction parts, which perform local and global learning, respectively. Nevertheless, this method only considers two tasks of image demosaicing and denoising. Ratnasingam \textit{et al.} reconstructed RGB images into RAW images to obtain a large number of training images~\cite{ratnasingam2019deep}. PyNET~\cite{ignatov2020replacing} focused on the mobile camera ISP pipeline and processed images from five different levels to obtain higher quality information. However, the processing time in the CPU mode is as high as 100 seconds.
 Moreover, some previous works~\cite{hsyu2021csanet,chaudhari2019merging,liang2021cameranet,kim2020pynet,dai2020awnet,raimundo2022lan,cheng2021lightweight} in AIM 2020 Challenge~\cite{ignatov2020aim,zhu2020eednet} and Mobile AI 2021 Challenge~\cite{ignatov2021learned} have achieved appealing results.
 CSANet~\cite{hsyu2021csanet} designed a double attention structure for mobile ISP while showing significant differences on other datasets. Moreover, the replacement of ISP is also based on HDR~\cite{chaudhari2019merging} and extreme low-light~\cite{zamir2021learning}. CameraNet~\cite{liang2021cameranet} defined ISP as restoration and enhancement subtasks, which are learned in two stages. More
 
 In terms of replacing an existing handcrafted ISP pipeline, these methods pay too much attention to the improvement of objective indicators or can only accomplish closely related tasks. In this paper, we abandon the redundant architecture to replace the model of the entire ISP pipeline. Moreover, we try to surpass the work of predecessors in ensuring a lightweight model and realizing the real coordinated development of ISP and AI vision.

\section{Proposing LW-ISP}
\label{base model}

\begin{figure*}[t]
\setlength{\abovecaptionskip}{0pt}
\setlength{\belowcaptionskip}{-15pt}
\centering
   \includegraphics[width=0.9\linewidth]{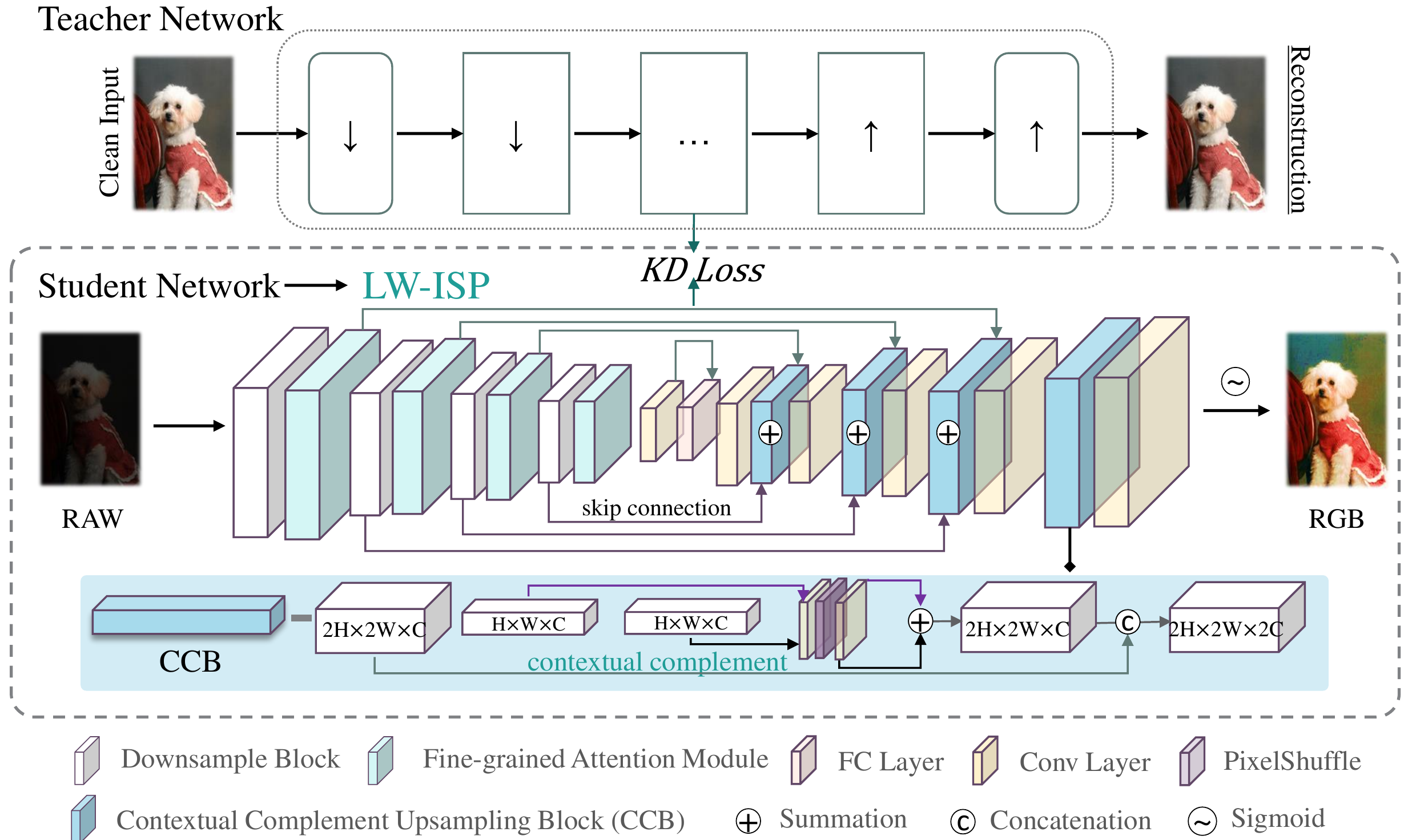}
   \caption{The overview of our method \texttt{LW-ISP}. The bottom half is the main architecture, which receives RAW input and execution feature reconstruction. 
   }
\label{overview}
\end{figure*}

 In this section, we present an overview of our proposed \texttt{LW-ISP}, as illustrated in Fig.~\ref{overview}. 
 Instead of naively adopting multi-scale or serial modular architecture to process RAW input, we take advantage of sophisticated structure and training strategies to achieve lightweight and high performance. We design a tiny U-Net as the backbone, which only contains 24 layers of convolution.
 The bottom half of Fig.~\ref{overview} shows our proposed end-to-end image preprocessing network, \texttt{LW-ISP}, which is composed of two stages. The first stage progressively downsamples feature maps at different levels to accelerate the computation. The second stage further concatenates the processed global vector with the tensor of the same size in the first half of the network through the symmetric skip connection. \textit{More details and codes about our architecture can be found in the supplementary material.}
 
  \vspace{-1.1em}


\subsection{Fine-Grained Attention Module}
\label{attention module}
\begin{figure}[t]
\centering
  \begin{minipage}[]{0.43\textwidth}
    \centering
    \includegraphics[width=1.0\linewidth]{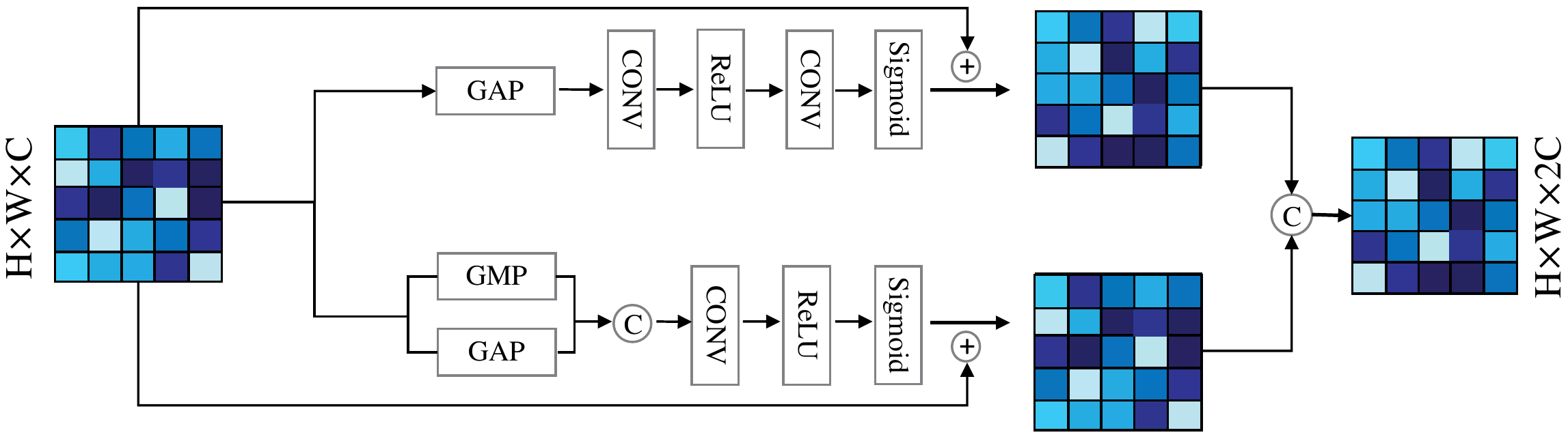}
    \caption{The fine-grained attention module (FGAM) combines channel attention (upper) and spatial attention (lower) with the input feature and then concatenate together. 
    }
    \label{attention}
  \end{minipage}
  \begin{minipage}[]{0.53\textwidth}
    \centering
    \includegraphics[width=1.0\linewidth]{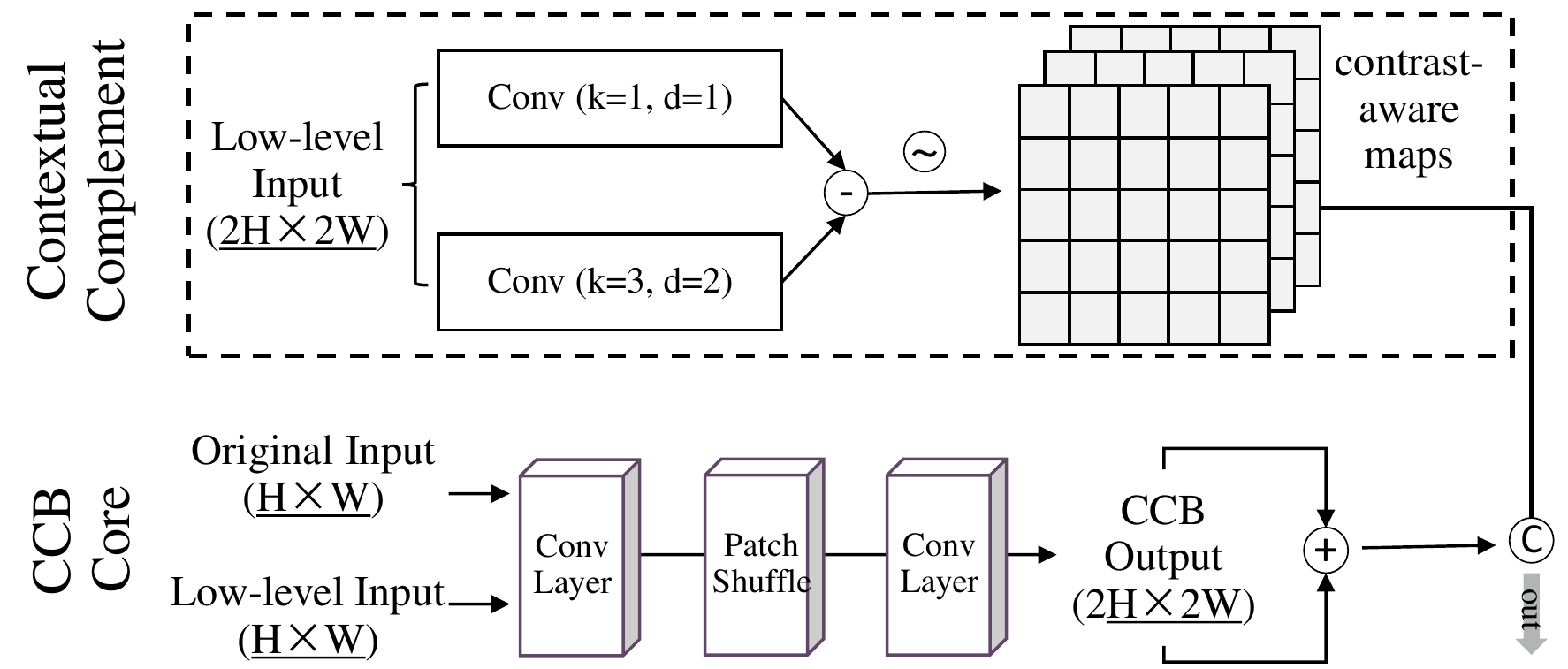}
    \caption{The contextual complement upsampling block (CCB). 
    }
    \label{3-CCB}
  \end{minipage}
\end{figure}

 As \texttt{LW-ISP} performs feature extraction on the RAW input in the first stage, more effective and discernible features need to be passed on. Existing attention fusion structures~\cite{woo2018cbam,hu2018squeeze} can reconstruct pixels in high-dimensional spaces. Furthermore, some research has been proposed to implement the attention mechanism in low-level vision but exhibited completely different negative effects in the ISP task~\cite{hong2020distilling,zamir2020learning}. \\
 To address this issue, we propose the \textit{Fine-grained Attention Module} (\textbf{FGAM}) suitable for ISP, as shown in Fig.~\ref{attention}. The feature recalibration is achieved by performing the attention of the channel and spatial dimensions in parallel and then indirectly fusing the intermediate features. As to channel attention, given a feature map $H\times W\times C$, the squeeze operation applies global average pooling (GAP) and two convolution layers followed by sigmoid gating to generate activation vector $1\times 1\times C$. As for spatial attention, GAP and max pooling (GMP) operate on feature map $H\times W\times C$ along the channel dimension and concatenate the outputs to yield a feature map $H\times W\times 2$. Then passing through a convolution layer and sigmoid to obtain the attention map $H\times W\times 1$. Specifically, in order to prevent the attention mechanism from suppressing the characteristics of irrelevant regions, we find that the direct fusion of spatial and channel attention should be avoided. In the FGAM, these two attention maps separately add to the input feature and then concatenate across channel dimension.

\subsection{Contextual Complement Upsampling Block}
\label{ccb}
The \textit{Contextual Complement Upsampling Block} (\textbf{CCB}) is composed of a contextual separation module for adaptive high frequency decomposition in the feature space, followed by the CCB-Core that fuses the corresponding size ($H \times W$) features of the previous stage. Unlike PyNET~\cite{ignatov2020replacing} that hand over the upsampling operation to the conventional method (bilinear interpolation) to solve, we switch to PixelShuffle~\cite{shi2016real} and design CCB-Core to suppress the loss of image information during zooming. \\
 \textbf{CCB-Core.} As shown by the dashed box in Fig.~\ref{overview} and Fig.~\ref{3-CCB}, in addition to the original features to be sampled, the input of CCB-Core also has the corresponding features ($2H\times2W$ for contextual decomposition and $H \times W$ for feature complementation) of the previous stage. The CCB-Core first performs a sub-pixel convolution on input features ($H \times W$ features from the first and second stages) to obtain global- and local-features. It should be noted that the convolution operations before and after PixelShuffle are used for channel dimension adjustment and fine-tuning, respectively. Subsequently, CCB-Core fuses the obtained features by residual learning to derive coarse high-resolution features. The operation of CCB-Core can be formulated as follows:
 \begin{equation}\mathcal{O}_{F A U}=\mathcal{P}\left(\mathcal{O}_{F B U}^{1}\right)+\mathcal{P}\left(\mathcal{O}_{F B U}^{2}\right),\end{equation}
 where $\mathcal{P}$ and $+$ stand for the functions of the sub-pixel convolution and residual learning, respectively. $FAU$/$FBU$ are short for the feature after/before upsampling, and the numbers represent the stage of the feature.\\
 \textbf{Contextual Complement.} This part is to learn contrast-aware features for image decomposition. To select adaptive contextual information, we first use two groups of dilated convolutions (with kernel size\&dilation rate of 1\&1 and 3\&2), denoted as $f_{d1}$ and $f_{d2}$, to extract features in different receptive fields. The effectiveness of this process will be verified in Section~\ref{ablation}. We then compute a contrast-aware map between the two feature maps as:
 \begin{equation}\mathcal{C}_{l}={sigmoid}\left(f_{d 1}\left(x_{i n}\right)-f_{d 2}\left(x_{i n}\right)\right),\end{equation}
 where $\mathcal{C}_l$ indicates the pixel-wise relative contrast information. $\mathcal{C}_l$ will eventually be concatenated to the $\mathcal{O}_{FAU}$ of CCB-Core to complete the contextual complement.

\subsection{Heterogeneous Distillation Algorithm}
\label{distillation}
 In knowledge distillation, although the source data distribution and processing dimensions of several networks may be different, target models can still be imitated by knowledge distillation on the target data~\cite{hou2019learning,hong2020distilling}. For ISP, its reconstruction process requires more hidden features and spatial structure information. In this paper, we propose a heterogeneous distillation algorithm, as shown in Fig.~\ref{overview}. The teacher will learn a wealth of intrinsic attributes from the clean inputs to assist the student network.
  
 For the teacher network, we continue to adopt the basic structure of \texttt{LW-ISP}. The only difference lies in the lack of an upsample block, due to the fact that RAW inputs will be half of the normal output after pre-processing. The teacher model will learn a clean mapping from the ground truth of the training data. The student model, \texttt{LW-ISP}, will be supervised during the training process, and no burden will be added during inference. We supervise the intermediate features and compute the feature similarity to control the imitation learning from the teacher to student. The determination of the position of the intermediate feature requires to be carefully selected. Besides, the heterogeneous distillation can also be used to extend the ISP pipeline to more fine-grained tasks.
 Experiments demonstrate that when applying the heterogeneous distillation, the position of the intermediate feature needs to be carefully selected, and the supervision provided in the first stage even harms the performance. We cherry-pick the output of the last three upsample blocks.

\subsection{Loss Function for LW-ISP}
\label{LW-ISP-loss}
 Based on the above distillation algorithm and full ISP's requirements for local and global correction and perceptual quality, we design multiple loss functions for training. The overall loss of \texttt{LW-ISP} can be formulated as:
 \begin{equation}\mathcal{L_{\text {overall}}}=\mathcal{L}_\text{{r}}+\alpha \cdot \mathcal{L}_\text{{s}}+\beta \cdot \mathcal{L}_\text{{d}},\end{equation}
 where $\alpha$ and $\beta$ are two hyper-parameters to balance the magnitude of $\mathcal{L}_s$ and $\mathcal{L}_d$. The sensitivity study and ablation study have been shown in Section~\ref{ablation}.
 
 \textbf{Reconstruction Loss.} Given a training sample $I$, the model result and ground truth can be denoted as $f(I)$ and $J$. To obtain the reconstruction result, we adopt the mean absolute error (MAE) to measure the difference:
 \begin{equation}\mathcal{L}_{r}=\left|f(I)-J\right|.\end{equation}
 
 \textbf{Structural Loss.} The multi-scale structural similarity (MS-SSIM~\cite{wang2003multiscale}) loss is used here to increase the dynamic range of the reconstructed photos:
 \begin{equation}\mathcal{L}_{s}=1-MS\cdot SSIM(f(I), J).\end{equation}

 \textbf{Distillation Loss.} The teacher network transmits the implicit information of the clean image to the student through the intermediate representations. Denote $F_m$ to be the feature maps of the $m_{th}$ layer of the model, the same is true for $n$. The distillation loss (KD loss) is formulated as:
 \begin{equation}\mathcal{L}_{d}=\sum_{(m, n) \in C} \mathcal{L}_{2}\left(F_{m}^{s}(I),\left(F_{n}^{t}(J)\right)\right),\end{equation}
 where $\mathcal{L}_2$ is the $L_2$-norm loss and $C$ is a set of candidate pairs of feature locations. In this work, $m$ and $n$ are set to be consistent. The superscript $t$ and $s$ denote the teacher model and student model, respectively.
 
 \subsection{Loss for Teacher Network}
 To learn an effective feature representation from the teacher model, we design the following loss function:
 \begin{equation}\mathcal{L}_{T}=(g(J)-J)^{2}+\gamma \cdot \mathcal{L}_{s},\end{equation} 
 where $g$ is the transform function and $J$ is the clean image. $\gamma$ is a hyper-parameter to balance the magnitude.

\section{Experiment}
 When learning RAW-to-RGB mapping with deep learning methods, we refer to it as \textbf{smart ISP}. If the smart ISP is to be applied towards landing application, the first thing to solve is the need for real-time inference and high imaging performance.
 In this section, we evaluate the effectiveness of our method on Zurich RAW to RGB~\cite{ignatov2020replacing} (Zurich for short) dataset, which is currently the largest ISP dataset. The effects of image denoising and enhancement are also evaluated on the SIDD~\cite{abdelhamed2018high}, DND~\cite{plotz2017benchmarking} and LoL~\cite{wei2018deep} datasets.

\begin{figure}[t]
\setlength{\abovecaptionskip}{0pt}
\setlength{\belowcaptionskip}{-10pt}
\centering
   \includegraphics[width=1.0\linewidth]{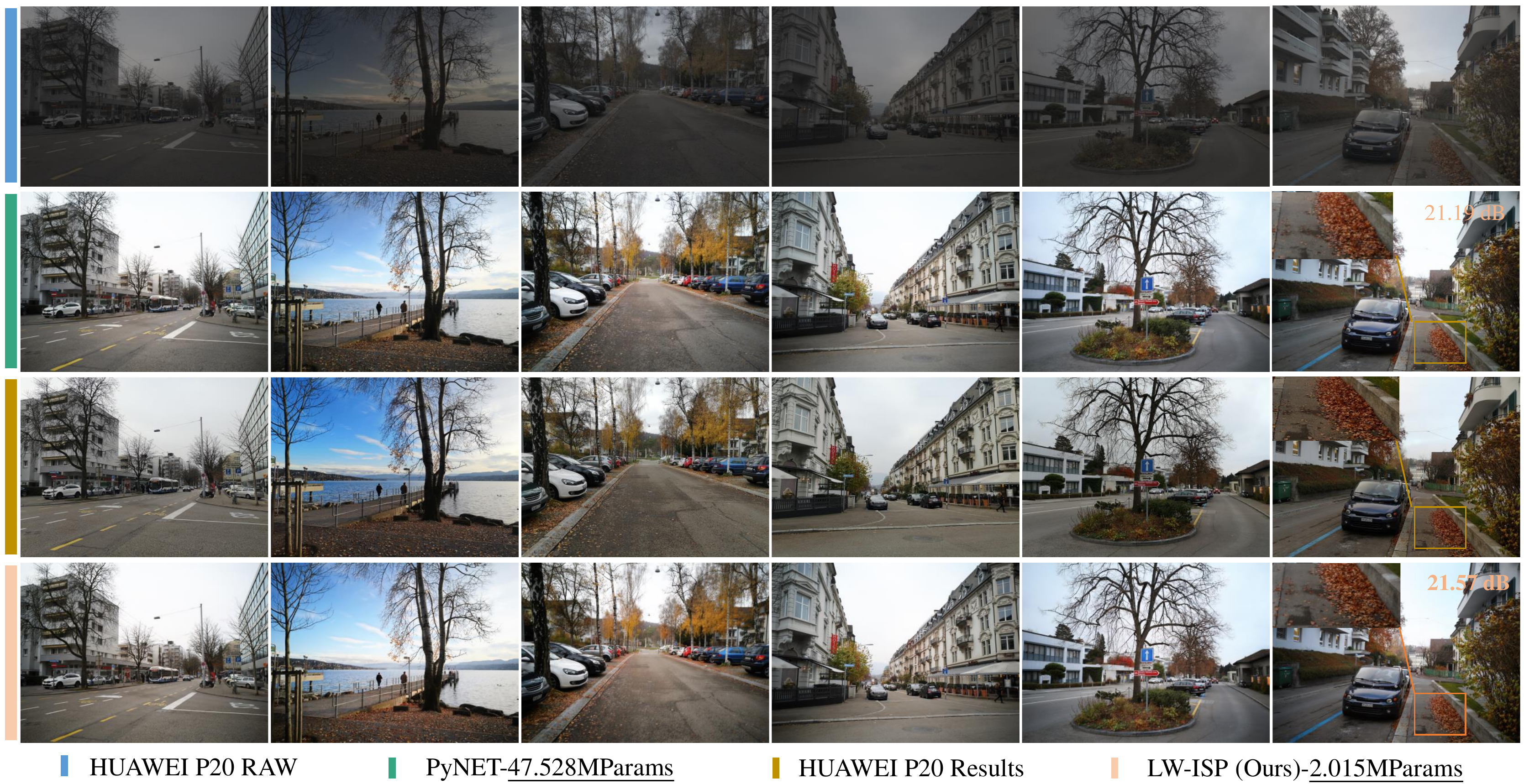}
   \caption{Sample visual results obtained with the proposed LW-ISP architecture (best zoomed on screen). These four lines represent the output of HUAWEI P20 sensor, state-of-the-art model PyNET, HUAWEI P20 camera and our method, respectively. Besides, the two numbers represent the evaluation results of the model with the PSNR.}
\label{qualitative figures}
\end{figure}

\begin{table}[t]
\centering
\resizebox{\textwidth}{!}{
\begin{tabular}{@{}c|ccccccccc@{}}
\toprule [1 pt]
Method  & SRCNN\cite{dong2015image}   & SRGAN\cite{ledig2017photo}  & DPED\cite{ignatov2017dslr}   & U-Net\cite{ronneberger2015u}  & Pix2Pix\cite{isola2017image} & SPADE\cite{park2019semantic}  & NAFNet\cite{chen2022simple} & PyNET\cite{ignatov2020replacing}           & LW-ISP      \\ \midrule
PSNR $(\uparrow)$    & 18.56  & 20.06  & 20.67  & 20.81  & 20.93   & 20.96 & 21.12 & \underline{21.19}           & \textbf{21.57} \\
MS-SSIM $(\uparrow)$ & 0.8268  & 0.8501 & 0.8560 & 0.8545 & 0.8532  & 0.8586 & 0.8613 & \underline{0.8620} & \textbf{0.8622}         \\ 
LPIPS $(\downarrow)$    & 0.385  & 0.257  & 0.343  & 0.257  & 0.208   & 0.209 & 0.194 & \underline{0.194}  & \textbf{0.160} \\
\bottomrule [1 pt]
\end{tabular}}
\caption{Comparison experiment results (PSNR/MS-SSIM/LPIPS) on Zurich Dataset \cite{ignatov2020replacing} (numbers in bold are the best). $\uparrow$ denotes that the upward trend corresponds to better performance, and $\downarrow$ denotes the downward trend.}
\label{table1}
\end{table}

\begin{table}[t]
\centering
\resizebox{0.95\linewidth}{!}{
\begin{tabular}{c|cccccc}
\toprule  [1 pt]
Model & Lightweight~\cite{cheng2021lightweight} & HERN~\cite{mei2019higher} & CameraNet~\cite{liang2021cameranet} & AWNet~\cite{dai2020awnet} & Pynet-ca~\cite{kim2020pynet} & LW-ISP (Ours)\\ \hline
PSNR (dB)  &  21.28   & 21.30  &  21.35    & 21.40   & 21.50 & \textbf{21.57}       \\
Params.(M) &  31.56 & 39.64 & 26.53 & 55.70 & 56.89 & \textbf{2.01}    \\ \bottomrule  [1 pt]
\end{tabular}}
\caption{\textit{Comparison with SOTA methods (PSNR and number of parameters) on Zurich Dataset \cite{ignatov2020replacing} (numbers in bold are the best).}}
\label{tablea}
\end{table}

 \subsection{Results on Zurich Dataset}
\label{zurich results}
 \textbf{Effectiveness of LW-ISP.} Due to limited deep learning research on the ISP pipeline, we compare different image preprocessing architectures, including the existing state-of-the-art method PyNET~\cite{ignatov2020replacing}. We adopt the mainstream image quality evaluation indicators PSNR, MS-SSIM and LPIPS. Table~\ref{table1} shows the quantitative performance of the proposed method on the real RAW to RGB mapping problem. It is obvious that LW-ISP outperforms other state-of-the-arts with the gain of at least 0.38dB in terms of PSNR and performs well on MS-SSIM. Fig.~\ref{demonstration} and Fig.~\ref{qualitative figures} compare the visual effects of LW-ISP with the previous best model PyNET and HUAWEI P20 after processing different RAW images. It is observed that the images taken by HUAWEI P20 are generally dark and over-render the sky and other backgrounds. 
 In contrast, the results of LW-ISP are more in line with the real characteristics and can generate better details, which are verified by objective metrics. The recently emerging NAFNet~\cite{chen2022simple} attempts to design a simple baseline for the field of image restoration, even removing nonlinear activation functions. Our method surpasses the state-of-the-art backbone (NAFNet) in low-level vision. We believe that the dark characteristics of the RAW data itself require more refined processing.
 
 \noindent
 \textbf{Efficiency of LW-ISP.} \textit{How much loss in quality is tolerable for the increase in speed?} Our model aims to achieve an effective balance between processing efficiency and algorithm performance to promote the development of the smart ISP. Roughly, we provide LW-ISP w/o FGAM and LW-ISP w/ FGAM in order to provide more options.
 LW-ISP w/o FGAM means that the attention module FGAM is not added to LW-ISP. As shown in Table~\ref{table-param}, compared to the SOTA PyNET~\cite{ignatov2020replacing}, LW-ISP can achieve up to 28.6 times of compression, and adding FGAM can still reduce it by 23 times. The results also present the floating point operations (FLOPs) under different resolution conditions. The calculation of the latest LW-ISP is 81 times less than PyNET. What is surprising is that LW-ISP can still achieve better performance than the previous best method with a minimum acceleration of 15 times. When testing on NVIDIA Tesla V100 GPU, LW-ISP takes 0.25 seconds to process 12MP photo (2944$\times$3958 pixels), while PyNET takes 3.8 seconds. \textit{The memory usage and other params of our model are presented in the supplementary.}
 
 \noindent
 \textbf{Comparison with SOTA Methods.} To solidify the performance of our method, we compare the performance of more recent methods. Based on fair comparisons, comparative experiments are performed without data augmentation and following our data preparation format (no extra input or pre-training). We show PSNR and the number of parameters in Table~\ref{tablea}.
 

\begin{table}[]
\resizebox{\linewidth}{!}{
\begin{tabular}{c|c|ccc|c}
\toprule [1 pt]
\multirow{2}{*}{Model} & \multirow{2}{*}{Number of Parameters} & \multicolumn{3}{c|}{FLOPS} & \multirow{2}{*}{PSNR (dB)} \\ \cline{3-5}
 &  & (224,224) & (960,960) & (1440,1984) &  \\ \midrule
SPADE~\cite{park2019semantic} & 97, 480, 899 & 191.31G & 3.16T & 10.89T & 20.96 \\
PyNET~\cite{ignatov2020replacing} & 47, 554, 738 & 342.698G & 5.72T & 19.513T & 21.19 \\
LW-ISP w/o FGAM & \textbf{1, 660, 777} & \textbf{3.441G} & \textbf{63.198G} & \textbf{195.914G} & \underline{21.40} \\
LW-ISP w/ FGAM  & \underline{2, 014, 681} & \underline{4.234G} & \underline{69.198G} & \underline{211.914G} & \textbf{21.57} \\ \bottomrule [1 pt]
\end{tabular}}
\caption{Comparison of the number of the parameters and FLOPs (floating point operations) between our proposed method and the state-of-the-art methods. Note that LW-ISP w/o FGAM does not adopt heterogeneous distillation and SPADE~\cite{park2019semantic} is a linear architecture with basically no long-distance cross-layer connection.}
\label{table-param}
\end{table}

\subsection{Results on Subtasks}
\label{subtask results}

In this section, we evaluate LW-ISP on subtasks to further explore the potential of end-to-end ISP model to reduce the calculation, comprehensively handle various tasks and generalize migration. Specifically, we demonstrate the effectiveness on denoising and enhancement.

\begin{table*}[t]
\resizebox{\textwidth}{!}{
\begin{tabular}{@{}c|ccccccccccc@{}}
\toprule [1 pt]
Method & DnCNN\cite{zhang2017beyond} & TNRD\cite{chen2015learning}   & BM3D\cite{dabov2007image}  & WNNM\cite{gu2014weighted}  & KSVD\cite{aharon2006k}  & EPLL\cite{zoran2011learning}  & CBDNet\cite{guo2019toward} & RIDNet\cite{anwar2019real}   & VDN\cite{yue2019variational}  & LW-ISP            & MIRNet\cite{zamir2020learning}                       \\ \midrule
PSNR   & 23.66 & 24.73 & 25.65 & 25.78 & 26.88 & 27.11 & 30.78  & 38.71  & 39.28                        & \textbf{39.44} & {\color[HTML]{6665CD} 39.72} \\
SSIM   & 0.583 & 0.643 & 0.685 & 0.809 & 0.842 & 0.870 & 0.754  & 0.914  &  0.909 & \textbf{0.918}  & {\color[HTML]{6665CD} 0.959} \\ \bottomrule [1 pt]
\end{tabular}}
\caption{Denoising results on the SIDD dataset~\cite{abdelhamed2018high}. Compared to the previous methods, our LW-ISP (numbers in bold) demonstrates a comparable performance. Blue font indicates the value higher than our method.}
\label{table4}
\end{table*}

 \noindent
 \textbf{Image Denoising.} We train our network only on the training set of SIDD and directly evaluate it on the test images of both SIDD and DND datasets. Quantitative comparisons in terms of PSNR and SSIM metrics are summarized in Table~\ref{table4} for SIDD. These experimental results show the excellent performance of our LW-ISP under such lightweight conditions, which has surpassed traditional data-driven algorithms. Furthermore, it is worth noting that our method provides better results while RIDNet~\cite{anwar2019real} uses additional training data, yet VIDNet~\cite{yue2019variational} and MIRNet~\cite{zamir2020learning} are much larger than LW-ISP.
 
 \noindent
 \textbf{Image Enhancement.} Without using any additional data and tricks, we achieve quite competitive results on the LoL~\cite{wei2018deep}. The PSNR can reach 20.18 dB, surpassing previous methods such as CRM~\cite{ying2017new} and MF~\cite{fu2016weighted}.
 

\subsection{Ablation Study}
\label{ablation}

\begin{figure}[t]
\centering
  \begin{minipage}[]{0.47\textwidth}
    \centering
    \includegraphics[width=1.0\linewidth]{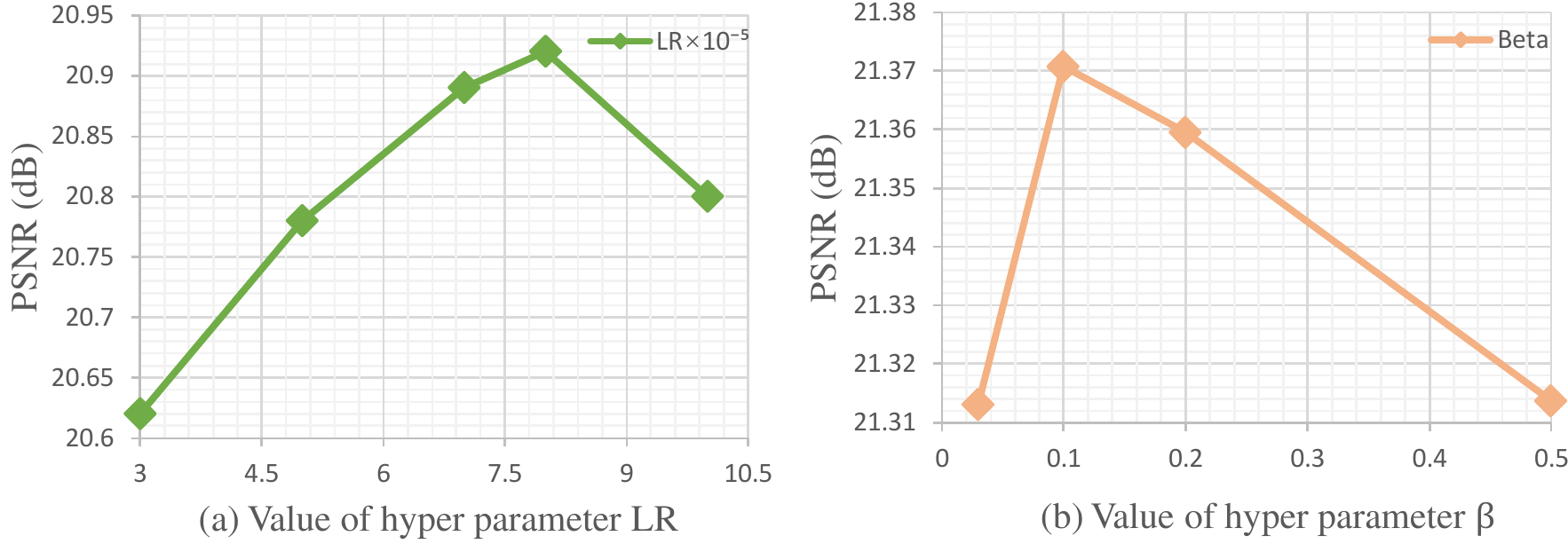}
    \caption{Sensitivity results of the parameters. (\textbf{a}) Learning Rate (LR): The initial learning rate in the training process. (\textbf{b}) Beta: The hyper-parameter $\beta$ in Equation (1) for the distillation loss $\mathcal{L}_d$.}
    \label{ablation params}
  \end{minipage}\quad
  \begin{minipage}[]{0.5\textwidth}
\resizebox{\linewidth}{!}{
\begin{tabular}{@{}ccccc@{}}
\toprule [1 pt]
Backbone & FGAM & CCB & Distillation & PSNR (dB)     \\ \midrule
$\checkmark$        &      &      &              & 20.92          \\
$\checkmark$        & $\checkmark$    &      &              & 21.17          \\
$\checkmark$        &      & $\checkmark$    &              & 21.28          \\
$\checkmark$        & $\checkmark$    & $\checkmark$    &              & 21.36          \\
$\checkmark$        &     & $\checkmark$    &  $\checkmark$    & 21.40         \\
$\checkmark$        & $\checkmark$    & $\checkmark$    & $\checkmark$            & \textbf{21.57} \\ \bottomrule [1 pt]
\end{tabular}}
  \tabcaption{Ablation study of our method. Backbone refers to the basic architecture of Section~\ref{base model}.}
\label{table5}
  \end{minipage}
\end{figure}

 \textbf{Study on Hyper-parameters.}
  In Fig.~\ref{ablation params} we show a sensitivity analysis of the parameters, which are used for LW-ISP training on the ISP pipeline and its subtasks. It can be observed that the LW-ISP backbone achieves the effect that we reported (20.92 dB) when the initial learning rate is $8\times10^{-5}$. In subfigure (b), we show the sensitivity study of $\beta$ during LW-ISP training (without structural loss $\mathcal{L}_s$). It can be easily observed that our performance boost holds for different values of $\beta$. The value of $\alpha$ has been set at 0.4.
  
 \noindent
 \textbf{Study on Architectural Components.}
 Table~\ref{table5} shows that FGAM and CCB can bring gains of 0.25 dB and 0.36 dB respectively, and the combination of the two can bring 0.44 dB gains. The baseline LW-ISP contains the base U-Net with a global feature vector and is trained with $\mathcal{L}_r+\mathcal{L}_s$, denoted as the backbone. Moreover, the implicit features from clean image learning by distillation are helpful and boost performance. \textit{More sensitivity studies of the hyper-parameters and ablation studies of FGAM can be found in the supplementary}.

 \noindent
 \textbf{Study on Distillation Algorithm.}
 When designing the distillation algorithm, a key insight is that we believe that the main structure (UNet) can be divided into image understanding and image reconstruction processes respectively according to the down/up sampling steps. (1) Our method can learn the reconstruction features corresponding to clean images under the supervision of the teacher during the upsampling reconstruction process. (2) We selected the outputs of the last three upsampling blocks. When performing ablation experiments at clean feature locations, the outputs of the four down/up sampling blocks are denoted as $down_i$ and $up_i$, respectively. It turns out that the closer the supervision to the RGB output location, the more performance gains (baseline: 21.28dB, $down_1+down_2+down_3$: 21.11dB, $down_4 +up_1+up_2$: 21.33dB, $up_2+up_3+up_4$: 21.40dB).

\noindent
\textbf{Study on FGAM.} 
 As to Fine-Grained Attention Module (FGAM), we use an addition operation (+) instead of multiplication (*) to generate channel attention and spatial attention. The reason is that we are surprised to find that the training process corresponding to the addition operation is more stable, and the performance is better (+21.17dB, *21.05dB). To perform a fair comparison, we experiment with the attention mechanism on the backbone. Specifically, channel attention and spatial attention respectively achieved 21.10 dB (+0.18) and 21.04 dB (+0.12). Our experimental results demonstrate that the combination fashion of channel attention and spatial attention in the FGAM is pretty essential. Unlike MIRNet~\cite{zamir2020learning} or CBAM~\cite{woo2018cbam}, the attention map output by each branch will be added to the input feature point-to-point, and then the two will be concatenated across channel dimension. It is observed that: \textbf{(a)} The multiplication of the attention map and the input feature that will lead to sub-optimal results (21.05dB). Note that the baseline and the results after adding FGAM are 20.97 dB and 20.17 dB, respectively. \textbf{(b)} It is unnecessary to add a convolutional layer with a kernel size of 1$\times$1 to reduce the number of channels. It will reach a result of 20.11 dB ($+$0.14 dB), still 0.06 dB from the optimal result. \textbf{(c)} The FGAM position should be placed after the downsample block. If the position is reversed, it only reaches 20.04 dB ($+$0.07 dB), which is 0.13 dB away from the optimal result.


\section{Discussion and Future Work}
\label{discussion}
\noindent
\textbf{Deep Application of ISP.}
 With the further collaborative development of ISP and AI vision, we believe that the collaboration between RAW data processed by neural networks and subsequent DL-based tasks will be more worth looking forward to in both theory and application. For instance, there is no unified conclusion on the objective evaluation standard of the processed photos. The effect of the smart ISP on the subsequent tasks may be a brand new viewpoint. What's more, smart ISPs can also integrate and learn from downstream tasks such as image recognition. The cooperation between the CNN accelerator and the ISP also requires a unique design. The pipeline that derives from this integration of hardware-friendly tradition and smart ISP will be a new direction.

\section{Conclusion}
 In this paper, we propose the \texttt{LW-ISP} to achieve real-time and high performance processing in smart ISP. The entire network implicitly learns the image mapping from RAW data to RGB photos, and utilizes the fine-grained attention module (FGAM), the contextual complement block (CCB) and heterogeneous distillation algorithm to reconstruct high-quality images. Abundant experiments show that \texttt{LW-ISP} has achieved state-of-the-art performance. 
 The model parameters have been reduced by 23$\times$ and the inference time has been accelerated by at least 15$\times$. Specifically, the results show that a U-Net (trained in the right manner) can replace much larger nets.
  

\clearpage
\bibliography{egbib}
\end{document}